\documentclass[10pt,twocolumn,letterpaper]{article}

\usepackage[pagenumbers]{cvpr}
\usepackage{tabularx}
\usepackage{multirow}
\usepackage{array}
\usepackage{enumitem}
\usepackage{microtype}

\newcolumntype{Y}{>{\raggedright\arraybackslash}X}
\newcommand{\papername}{this survey}

\setlist[itemize]{leftmargin=*,topsep=2pt,itemsep=2pt,parsep=0pt}
\definecolor{cvprblue}{rgb}{0.21,0.49,0.74}
\usepackage[pagebackref,breaklinks,colorlinks,allcolors=cvprblue]{hyperref}

\def\confName{CVPR}
\def\confYear{2026}

\title{Physical Adversarial Attacks on AI Surveillance Systems:\\
Detection, Tracking, and Visible--Infrared Evasion}

\author{
Miguel A. Dela Cruz\\
Department of Computer Science\\
University of the Philippines Diliman\\
Quezon City, Philippines
\and
Patricia Mae Santos\\
Department of Information Systems and Computer Science\\
Ateneo de Manila University\\
Quezon City, Philippines
\and
Rafael T. Navarro\\
College of Computer Studies\\
De La Salle University\\
Manila, Philippines
}

\begin{document}
\maketitle

\begin{abstract}
Physical adversarial attacks are increasingly studied in settings that resemble deployed surveillance systems rather than isolated image benchmarks. In these settings, person detection, multi-object tracking, visible--infrared sensing, and the practical form of the attack carrier all matter at once. This changes how the literature should be read. A perturbation that suppresses a detector in one frame may have limited practical effect if identity is recovered over time; an RGB-only result may say little about night-time systems that rely on visible and thermal inputs together; and a conspicuous patch can imply a different threat model from a wearable or selectively activated carrier. This paper reviews physical attacks from that surveillance-oriented viewpoint. Rather than attempting a complete catalogue of all physical attacks in computer vision, we focus on the technical questions that become central in surveillance: temporal persistence, sensing modality, carrier realism, and system-level objective. We organize prior work through a four-part taxonomy and discuss how recent results on multi-object tracking, dual-modal visible--infrared evasion, and controllable clothing reflect a broader change in the field. We also summarize evaluation practices and unresolved gaps, including distance robustness, camera-pipeline variation, identity-level metrics, and activation-aware testing. The resulting picture is that surveillance robustness cannot be judged reliably from isolated per-frame benchmarks alone; it has to be examined as a system problem unfolding over time, across sensors, and under realistic physical deployment constraints.
\end{abstract}

\section{Introduction}
\label{sec:intro}

Adversarial robustness in computer vision was first discussed largely in terms of classifiers and, later, standalone detectors, often under digitally generated perturbations \cite{akhtar2018threat,akhtar2021advances}. Physical attacks changed that picture by forcing perturbations to survive printing, viewpoint changes, lighting variation, and camera processing. Once the problem is placed in a surveillance setting, the difference between benchmark-style success and operational impact becomes even clearer.

A surveillance pipeline is rarely just a detector applied to isolated frames. It typically combines detection, tracking, identity association, and, in many low-light deployments, multiple sensing channels such as visible and thermal cameras. Under those conditions, the key question is not only whether a target disappears in one image. An attack may instead matter because it survives across time, induces identity confusion, exploits disagreements between sensing channels, or can be deployed as a garment or accessory without making the threat model implausible.

Several surveys provide the necessary background. Broad reviews cover adversarial attacks and defenses across tasks and modalities \cite{akhtar2018threat,akhtar2021advances}. Later surveys concentrate on the physical world, emphasizing carriers, optimization procedures, and real-world constraints \cite{wang2022physicalsurvey,wei2022decade,wang2023physicalworldsurvey}. More recent work has begun to discuss surveillance and optical attacks directly \cite{nguyen2023surveillance,fang2024optical}. Together, these papers make clear that surveillance is not just another application area. It is a setting in which time, sensing redundancy, and deployment realism interact.

This survey takes that narrower viewpoint. We focus on physical attacks against AI surveillance systems used for person detection, tracking, and low-light monitoring, and we organize the literature around four practical questions: what persists over time, which sensors are involved, what carrier is physically deployed, and what failure is induced at the system level. Recent results on multi-object tracking \cite{long2024papmot}, dual-modal visible--infrared attacks \cite{long2025cdupatch}, and thermally activated clothing \cite{long2026thermally} are especially informative in this respect because they make surveillance-specific evaluation requirements hard to ignore.

Our goal is not to catalogue every physical attack in computer vision. Instead, we use a surveillance-centered taxonomy to read the literature in a way that foregrounds persistent identity disruption, visible--infrared evasion, and wearable or controllable carriers. We then examine what those shifts imply for experimental protocols, defensive design, and benchmark construction.

The remainder of the paper is organized as follows. \Cref{sec:related} positions our scope relative to existing surveys. \Cref{sec:taxonomy} introduces the taxonomy used throughout the paper. \Cref{sec:transitions} discusses how the surveillance threat model has broadened over time. \Cref{sec:evaluation} summarizes evaluation practices, defensive implications, and open problems. \Cref{sec:conclusion} concludes.

\section{Related Work and Scope}
\label{sec:related}

Physical attacks on surveillance systems sit at the intersection of several survey literatures. General adversarial-vision surveys provide the broadest background across tasks and defenses \cite{akhtar2018threat,akhtar2021advances}. Surveys dedicated to the physical world narrow the focus to printability, viewpoint robustness, carrier design, and experimental realism \cite{wang2022physicalsurvey,wei2022decade,wang2023physicalworldsurvey}. More recent reviews have begun to discuss surveillance applications explicitly \cite{nguyen2023surveillance}, while optical physical attacks have been treated as their own sub-area because illumination, projection, reflection, and sensor effects raise different challenges from those of printed patches alone \cite{fang2024optical}.

What remains less developed is a view that treats the surveillance pipeline itself as the main object of study. Broad surveys necessarily compress the differences between one-frame detection, long-horizon tracking, and visible--infrared monitoring. Surveys organized mainly by carrier type can likewise obscure the fact that the same physical artifact may have very different consequences depending on whether the target system performs detection, association, or cross-modal fusion.

This is where our scope differs. We are interested in physical attacks that matter for persistent monitoring systems: person detection in public scenes, tracking-by-detection pipelines, and visible--infrared deployments used in low-light or night-time settings. Recent papers on physical attacks against multi-object tracking, universal dual-modal patches, and thermally activated clothing are part of that story \cite{long2024papmot,long2025cdupatch,long2026thermally}, but they are considered here as representative signs of a broader shift rather than as the whole field.

\begin{table*}[t]
\centering
\scriptsize
\caption{Positioning this paper relative to existing surveys. ``Surveillance specificity'' asks whether persistent monitoring pipelines are treated as a primary object of study rather than a side application.}
\label{tab:surveys}
\setlength{\tabcolsep}{3pt}
\begin{tabularx}{\textwidth}{@{}p{3.2cm}p{1.0cm}p{2.2cm}p{1.5cm}p{1.2cm}p{1.2cm}Y@{}}
\toprule
Survey or perspective & Year & Primary scope & Surveillance specificity & Tracking emphasis & VI/IR emphasis & Distinctive value relative to this paper \\
\midrule
Akhtar and Mian \cite{akhtar2018threat} & 2018 & General adversarial vision survey & Low & Low & Low & Foundational but predates the recent wave of surveillance-focused physical attacks. \\
Akhtar \etal \cite{akhtar2021advances} & 2021 & General attack/defense survey & Low & Low & Low & Broader than our scope and not centered on physical surveillance pipelines. \\
Wang \etal \cite{wang2022physicalsurvey} & 2022 & Physical attacks in computer vision & Medium & Low & Medium & Strong taxonomy of carriers and tasks, but less focused on surveillance operations. \\
Wei \etal \cite{wei2022decade} & 2022 & Ten-year review of physical attacks & Medium & Low & Medium & Historical breadth; less emphasis on recent surveillance-specific transitions. \\
Wang \etal \cite{wang2023physicalworldsurvey} & 2023 & Physical adversarial examples in the real world & Medium & Low & Medium & Covers the physical-world bridge well, but not the surveillance pipeline in depth. \\
Nguyen \etal \cite{nguyen2023surveillance} & 2023 & Surveillance-focused physical attacks & High & Medium & Medium & Closest in spirit; our paper sharpens the perspective around tracking, VI sensing, and wearability. \\
Fang \etal \cite{fang2024optical} & 2024 & Optical physical attacks & Medium & Low & Low & Useful for optical carriers, but not targeted at surveillance persistence or dual-modal sensing. \\
This paper & 2026 & Surveillance-centered review & High & High & High & Brings together tracking, visible--infrared sensing, and carrier realism within a single surveillance-oriented scope. \\
\bottomrule
\end{tabularx}
\end{table*}

We also depart from views that separate papers only by carrier or only by target task. In surveillance, these axes interact. A sticker, a T-shirt, an infrared block, and thermally manipulated clothing are different carriers, but their significance depends strongly on whether the target is a one-frame detector, a multi-object tracker, or a dual-modal detector. Likewise, tracking and detection attacks may share infrastructure while leading to different failure modes: missed detections can suppress an alarm, whereas tracker corruption can generate ID switches or long-lived false trajectories \cite{luo2021motreview,deepsort2017}.

Our scope is deliberately technical rather than policy-led. AI surveillance also raises social and governance questions, but the purpose of \papername\ is to clarify technical threat models, evaluation gaps, and design principles for physical attacks and defenses. Within that frame, the next section introduces a surveillance-centered taxonomy for organizing the literature more concretely.

\section{A Surveillance-Centered Taxonomy}
\label{sec:taxonomy}

Surveillance systems differ from generic vision benchmarks in one critical way: the model is embedded in an operational loop. A detector may trigger recording or alerting; a tracker may accumulate identity evidence over dozens of frames; visible and thermal cameras may be fused before decision making; and the physical carrier of an attack must coexist with clothing, scene geometry, weather, and camera processing. This motivates a taxonomy that emphasizes system behavior rather than only optimization technique.

\begin{figure*}[t]
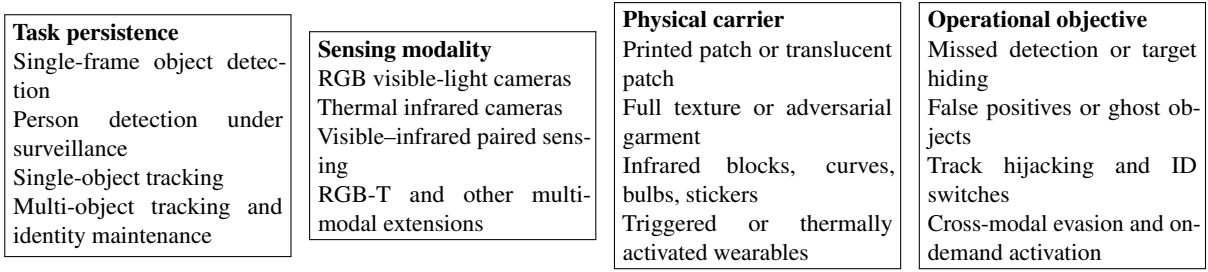

\centering
\small
\setlength{\tabcolsep}{3pt}
\begin{tabular}{p{0.22\textwidth}p{0.22\textwidth}p{0.22\textwidth}p{0.22\textwidth}}
\fbox{\parbox{0.205\textwidth}{\textbf{Task persistence}\\
Single-frame object detection\\
Person detection under surveillance\\
Single-object tracking\\
Multi-object tracking and identity maintenance}} &
\fbox{\parbox{0.205\textwidth}{\textbf{Sensing modality}\\
RGB visible-light cameras\\
Thermal infrared cameras\\
Visible--infrared paired sensing\\
RGB-T and other multi-modal extensions}} &
\fbox{\parbox{0.205\textwidth}{\textbf{Physical carrier}\\
Printed patch or translucent patch\\
Full texture or adversarial garment\\
Infrared blocks, curves, bulbs, stickers\\
Triggered or thermally activated wearables}} &
\fbox{\parbox{0.205\textwidth}{\textbf{Operational objective}\\
Missed detection or target hiding\\
False positives or ghost objects\\
Track hijacking and ID switches\\
Cross-modal evasion and on-demand activation}}
\end{tabular}
\caption{A surveillance-centered taxonomy of physical adversarial attacks. The point is not only to classify papers, but to clarify which operational questions a method actually stresses.}
\label{fig:taxonomy}
\end{figure*}

We organize the literature around four dimensions shown in \Cref{fig:taxonomy}.

\subsection{Task persistence}

The first dimension is whether the system acts on isolated frames or on temporally extended evidence. Classical object-detection attacks often measure success frame by frame \cite{xie2017dag,liu2018dpatch,chen2018shapeshifter}. Surveillance-specific person-detection work moves closer to deployment, especially when wearability and camera variation are modeled explicitly \cite{thys2019surveillance,xu2020tshirt,tan2021legitimate,hu2022advtexture,guesmi2024dap,wei2024cameraagnostic,cheng2024fulldistance}. Tracking attacks go further by making temporal persistence central: the perturbation must not merely suppress a detection once, but corrupt the trajectory or identity state over time \cite{wiyatno2019texture,chen2021multiscenario,jia2021iouattack,ding2021physicaltracking,lin2021trackletswitch,zhou2023ffattack,long2024papmot,pang2024blurmot}.

\subsection{Sensing modality}

The second dimension is sensing modality. RGB attacks remain the dominant baseline, but surveillance systems increasingly use thermal cameras at night or visible--infrared paired sensing in low-light environments. This introduces genuinely new constraints because the perturbation must either transfer across modalities or exploit thermal-specific carriers such as heating elements and infrared emitters \cite{zhu2021smallbulbs,zhu2022infraredclothing,wei2024irpatches,wei2023hotcold,zhu2023irclothes,hu2024irblocks,hu2024ircurves,zhu2024thermalhiding,zhu2024irstickers}. Cross-modal visible--infrared attacks make this explicit by optimizing for both channels at once rather than treating infrared as an afterthought \cite{tpamipatch,long2025cdupatch,long2026thermally}.

\subsection{Physical carrier}

The third dimension is the physical carrier. Printed patches remain the canonical starting point \cite{brown2017patch,liu2018dpatch}, but surveillance-relevant carriers quickly become more diverse: large patches for person detection \cite{thys2019surveillance}, full-body textures and adversarial garments \cite{xu2020tshirt,hu2022advtexture,zhu2023irclothes}, translucent or naturalistic patches that better fit scene appearance \cite{hu2021npap,zolfi2021translucent}, and thermal carriers such as bulbs, blocks, curves, stickers, and heated wearables \cite{zhu2021smallbulbs,wei2023hotcold,hu2024irblocks,hu2024ircurves,zhu2024irstickers}. Triggered and controllable carriers add another operational layer because the attacker can decide when the perturbation becomes active \cite{zhu2023tpatch,long2026thermally}.

\subsection{Operational objective}

The fourth dimension is the attack objective as it matters to surveillance operators. Some attacks hide a person or object \cite{thys2019surveillance,wu2020cloak}. Others create false positives or manipulate post-processing, which can flood systems with ghost detections or distract operators \cite{wang2019daedalus,zhou2023ffattack}. Tracking attacks further target identity integrity, producing ID switches or false track continuity \cite{lin2021trackletswitch,long2024papmot}. Cross-modal attacks seek simultaneous evasion across visible and thermal channels \cite{tpamipatch,long2025cdupatch}. Wearable and thermally activated attacks add the ability to look benign most of the time while activating only under chosen circumstances \cite{zhu2023tpatch,long2026thermally}.

\begin{table*}[t]
\centering
\scriptsize
\caption{Representative literature through the surveillance-centered taxonomy. We intentionally group by operational question rather than by venue or year.}
\label{tab:representative}
\setlength{\tabcolsep}{4pt}
\begin{tabularx}{\textwidth}{@{}p{2.0cm}Yp{2.2cm}Y@{}}
\toprule
Category & Representative works & Common carriers & Core surveillance question \\
\midrule
General patch foundations & \cite{brown2017patch,athalye2018synthesizing,eykholt2018robust,chen2018shapeshifter,liu2018dpatch,zhao2018seeing,wang2019daedalus} & Printed stickers, robust EoT patches, scene-integrated perturbations & How can attacks survive viewpoint change, printing, and detector-specific structure? \\
Person detection under surveillance & \cite{thys2019surveillance,xu2020tshirt,hu2021npap,tan2021legitimate,hu2022advtexture,guesmi2024dap,wei2024cameraagnostic,cheng2024fulldistance,zhu2023tpatch} & Wearable patches, garments, naturalistic textures, dynamic or triggered patches & How can attacks remain effective across distance, camera pipelines, and human-compatible deployment? \\
Tracking and temporal persistence & \cite{wiyatno2019texture,chen2021multiscenario,jia2021iouattack,ding2021physicaltracking,lin2021trackletswitch,zhou2023ffattack,pang2024blurmot,long2024papmot} & Textures, query-based perturbations, trajectory-level manipulations, printable patches & How does attack success accumulate over time through association, memory, and identity maintenance? \\
Thermal and infrared evasion & \cite{zhu2021smallbulbs,zhu2022infraredclothing,wei2024irpatches,wei2023hotcold,zhu2023irclothes,hu2024irblocks,hu2024ircurves,zhu2024thermalhiding,zhu2024irstickers} & Heated bulbs, infrared clothing, learnable IR patches, blocks, curves, stickers & Which carriers can survive thermal imaging constraints and real night-time deployment? \\
Visible--infrared cross-modal attacks & \cite{jia2021llvip,tpamipatch,long2025cdupatch,long2026thermally} & Unified VI patch, dual-modal clothing, cross-channel physical designs & What does it take to evade both visible and thermal channels simultaneously? \\
\bottomrule
\end{tabularx}
\end{table*}

This taxonomy matters because it changes how we interpret the literature. A method that is impressive as an RGB detection attack may still be weak as a surveillance attack if it fails under tracking, thermal sensing, or realistic wearability constraints. Conversely, a modest-looking attack can be more operationally serious if it induces persistent ID corruption or cross-modal evasion. The next section uses this lens to discuss how the surveillance threat model has broadened over time.

\section{Shifts in the Surveillance Threat Model}
\label{sec:transitions}

Read chronologically, the literature suggests that physical attacks on surveillance systems have broadened along three connected axes. Early work mostly asked whether a detector could be fooled in the wild. More recent papers ask harder questions: whether the error persists across time, whether it survives when visible and thermal signals are used together, and whether the attack can be carried in a wearable or selectively activated form. Older detector attacks remain important, but they no longer capture the full surveillance threat model on their own.

\subsection{From detector evasion to identity-aware persistence}

Much of the early physical-attack literature in surveillance was detector-centric. Person detectors were attacked with conspicuous but effective patches \cite{thys2019surveillance}, then with T-shirts, naturalistic textures, and visually plausible garments \cite{xu2020tshirt,hu2022advtexture,tan2021legitimate}. Later papers improved realism by examining camera variation, distance, and scene integration \cite{hu2021npap,guesmi2024dap,wei2024cameraagnostic,cheng2024fulldistance}. This line of work remains important because person detection is often the first stage of an automated surveillance stack.

Tracking research exposed what detector-only evaluation leaves out. Physical and digital attacks against tracking showed that the relevant failure mode is often not one-frame disappearance but long-lived drift, response corruption, or identity mistakes \cite{wiyatno2019texture,chen2021multiscenario,jia2021iouattack,ding2021physicaltracking}. Multi-object tracking makes the threat richer still because attacks can target data association itself. Tracklet-Switch and F\&F Attack already suggest that surveillance risk is tied to identity corruption, not just missed boxes \cite{lin2021trackletswitch,zhou2023ffattack}, and Pang \etal show that multi-object trackers can be both blinded and distracted by adversarial perturbations \cite{pang2024blurmot}.

Recent physical attacks on multi-object tracking make this shift explicit \cite{long2024papmot}. They move the evaluation target from frame-level detection accuracy to identity continuity over time, which is often closer to the failure mode that matters in practice. This perspective also aligns with how modern tracking-by-detection systems are deployed \cite{deepsort2017,luo2021motreview}: the key question is not only whether the camera sees a person, but whether the system keeps the right trajectory attached to the right identity.

\subsection{From RGB attacks to visible--infrared evasion}

The second shift comes from sensing modality. RGB cameras remain the default benchmark, but they are not the whole surveillance stack. Night-time monitoring, perimeter security, and low-light traffic scenarios often rely on thermal cameras or visible--infrared pairs, which makes purely visual camouflage less informative. Thermal attacks therefore developed their own design language, including bulbs, infrared clothing, learnable patches, and HOTCOLD-style wearable elements \cite{zhu2021smallbulbs,zhu2022infraredclothing,wei2024irpatches,wei2023hotcold,zhu2023irclothes,hu2024irblocks,hu2024ircurves,zhu2024thermalhiding,zhu2024irstickers}.

Once visible and infrared channels are considered together, the problem changes again. A method that succeeds in RGB alone may say little about a system that fuses visible and thermal information. Unified visible--infrared patches established that cross-modal physical attacks are feasible \cite{tpamipatch}. More recent dual-modal work pushes toward stronger transfer across detectors and scenes \cite{long2025cdupatch}. The broader lesson is that surveillance robustness should be measured against coupled sensing channels rather than against each modality in isolation.

This shift also affects what counts as a plausible carrier. In RGB benchmarks, one can often discuss visual inconspicuousness. In dual-modal settings, however, the carrier has to negotiate two appearance spaces with different physics. Visible--infrared evasion is therefore less like ordinary texture design and more like coordinated sensor-facing engineering.

\subsection{From static patches to wearable and controllable threats}

The third shift concerns deployment realism. A printed patch on a board or a fixed sticker on an object is an instructive starting point, but many surveillance-relevant attacks must live on clothes, accessories, or thermally manipulated materials. Person-detection papers already moved in this direction through adversarial garments and naturalistic textures \cite{xu2020tshirt,hu2022advtexture,tan2021legitimate}. Benchmark work such as REAP and recent camera-aware studies further emphasized that success under one camera or one distance is not enough to support strong claims about real use \cite{hingun2023reap,wei2024cameraagnostic,guesmi2024dap,cheng2024fulldistance}.

Controllability pushes the threat model further. Triggered patches show that an attack need not be permanently active \cite{zhu2023tpatch}. Thermally activated dual-modal clothing extends the same idea to visible--infrared surveillance \cite{long2026thermally}, showing that wearability, cross-modal evasion, and user control can be combined in a single carrier. This matters because the timing of activation changes both the attacker's options and the defender's assumptions. A system that appears stable during passive observation may still fail once the carrier is deliberately switched into an effective state.

\begin{figure*}[t]
\centering
\small
\fbox{%
\parbox{0.965\textwidth}{%
\textbf{2017--2018: foundations.} Universal patches, EoT-style robust examples, and detector-specific attacks establish the feasibility of physical attacks on modern vision systems \cite{brown2017patch,athalye2018synthesizing,eykholt2018robust,chen2018shapeshifter,liu2018dpatch,zhao2018seeing,wang2019daedalus}.\\[0.3em]
\textbf{2019--2020: surveillance entry points.} Person-detection patches and physical tracking textures bring physical attacks into surveillance-like settings \cite{thys2019surveillance,wiyatno2019texture,wu2020cloak,xu2020tshirt}.\\[0.3em]
\textbf{2021--2023: infrared and identity-aware attacks.} Thermal carriers, naturalistic garments, tracking attacks, and VI unified patches expand the threat model toward sensing diversity and temporal association \cite{zhu2021smallbulbs,zhu2022infraredclothing,wei2024irpatches,wei2023hotcold,chen2021multiscenario,jia2021iouattack,ding2021physicaltracking,lin2021trackletswitch,zhou2023ffattack,tpamipatch}.\\[0.3em]
\textbf{2024--2026: operational realism.} Camera-aware attacks, full-distance evaluation, multi-view infrared attacks, multi-object tracking attacks, stronger dual-modal patches, and thermally activated clothing make persistence, cross-modality, and controllability central evaluation axes \cite{wei2024cameraagnostic,hu2024irblocks,hu2024ircurves,zhu2024thermalhiding,zhu2024irstickers,guesmi2024dap,cheng2024fulldistance,long2024papmot,long2025cdupatch,long2026thermally}.%
}}
\caption{A timeline view of the literature. The field has moved from detector-centric demonstrations toward temporally persistent, cross-modal, and wearable attacks that better match surveillance deployments.}
\label{fig:timeline}
\end{figure*}

\begin{table*}[t]
\centering
\scriptsize
\caption{Selected milestones that broadened the surveillance threat model. The table mixes earlier external work with more recent papers that highlight tracking, dual-modal sensing, and controllable wearables.}
\label{tab:milestones}
\setlength{\tabcolsep}{4pt}
\begin{tabularx}{\textwidth}{@{}p{3.1cm}p{1.0cm}p{2.0cm}p{2.0cm}p{2.1cm}Y@{}}
\toprule
Representative work & Year & Primary task & Sensing modality & Carrier type & Why it mattered for surveillance \\
\midrule
Thys \etal \cite{thys2019surveillance} & 2019 & Person detection & RGB & Wearable adversarial patch & Established physical attacks directly against automated surveillance cameras. \\
Wiyatno and Xu \cite{wiyatno2019texture} & 2019 & Visual object tracking & RGB video & Adversarial texture & Showed that temporal objectives in tracking differ from one-frame detector evasion. \\
Wei \etal \cite{tpamipatch} & 2023 & Cross-modal detection & Visible--infrared & Unified physical patch & Demonstrated joint visible--infrared attacks in the physical world. \\
Wei \etal \cite{wei2023hotcold} & 2023 & Thermal pedestrian detection & Infrared & Wearable HOTCOLD block & Highlighted practical thermal carriers for human-centered evasion. \\
PapMOT \cite{long2024papmot} & 2024 & Multi-object tracking & RGB surveillance video & Printable patch & Brought physical patch attacks to identity association in multi-object tracking. \\
Cdupatch \cite{long2025cdupatch} & 2025 & Dual-modal detection & Visible--infrared & Universal dual-modal patch & Emphasized detector transfer and scene-level universality across two sensing channels. \\
Thermally Activated Dual-Modal Adversarial Clothing \cite{long2026thermally} & 2026 & Dual-modal surveillance evasion & Visible--infrared & Wearable, thermally activated clothing & Combined wearability and on-demand activation in a dual-modal surveillance setting. \\
\bottomrule
\end{tabularx}
\end{table*}

Taken together, these developments suggest that surveillance papers are most informative when they report how an attack behaves across time, across sensors, and under realistic carrying conditions. Per-frame RGB success remains a useful baseline, but it is no longer sufficient to characterize operational risk.

\section{Evaluation Protocols, Defenses, and Open Problems}
\label{sec:evaluation}

As the threat model broadens, evaluation has to do more than report per-frame attack success. The literature already points toward this change, but protocols remain uneven. What matters in surveillance is not only whether a detector fails once, but whether failure persists over time, survives across sensing channels, and remains plausible under real deployment constraints.

\subsection{What surveillance-grade evaluation should report}

Several ingredients recur across the most convincing papers. One is \emph{temporal persistence}: tracking attacks and surveillance deployments depend on what happens across frames, not just within them \cite{wiyatno2019texture,lin2021trackletswitch,zhou2023ffattack,long2024papmot}. Another is \emph{cross-modal consistency}: visible and thermal signals are often used together, so an RGB-only result may misstate practical risk in either direction \cite{jia2021llvip,tpamipatch,hu2024irblocks,hu2024ircurves,long2025cdupatch,long2026thermally}. Physical performance is also highly sensitive to camera and distance variation, as emphasized by REAP, DAP, camera-agnostic attacks, and full-distance evaluation \cite{hingun2023reap,guesmi2024dap,wei2024cameraagnostic,cheng2024fulldistance}. Finally, \emph{carrier realism} matters. A large printed patch, a visually plausible garment, and a selectively activated wearable each imply a different deployment story and a different defensive burden \cite{thys2019surveillance,hu2022advtexture,tan2021legitimate,zhu2023tpatch,long2026thermally}.

\begin{figure*}[t]
\centering
\scriptsize
\setlength{\tabcolsep}{2pt}
\resizebox{\textwidth}{!}{
\begin{tabular}{p{0.158\textwidth}p{0.158\textwidth}p{0.158\textwidth}p{0.158\textwidth}p{0.158\textwidth}}
\fbox{\parbox{0.148\textwidth}{\textbf{Stage 1}\\Digital tests\\Single model\\Per-frame metrics}} &
\fbox{\parbox{0.148\textwidth}{\textbf{Stage 2}\\Lab physical tests\\Printed or wearable carrier\\Short-range views}} &
\fbox{\parbox{0.148\textwidth}{\textbf{Stage 3}\\Operational variation\\Distance change\\Camera and ISP shifts}} &
\fbox{\parbox{0.148\textwidth}{\textbf{Stage 4}\\Temporal persistence\\Tracking and identity metrics\\Multi-frame success}} &
\fbox{\parbox{0.148\textwidth}{\textbf{Stage 5}\\Multimodal risk\\Visible + thermal\\On-demand activation}}
\end{tabular}
}
\caption{A five-stage evaluation ladder for surveillance-relevant physical attacks. Most papers cover the early stages; far fewer cover all five.}
\label{fig:evaluation-ladder}
\end{figure*}

\subsection{Why defenses remain fragmented}

Compared with the attack literature, surveillance-specific defenses are still fragmented. Broad surveys and benchmarks already suggest why: a defense that works against one carrier or one sensor often fails to generalize across cameras, distances, or modalities \cite{akhtar2021advances,wang2022physicalsurvey,nguyen2023surveillance,fang2024optical}. REAP and camera-agnostic studies argue that robustness claims should not be trusted without camera diversity \cite{hingun2023reap,wei2024cameraagnostic}. Tracking papers suggest that a defense also has to inspect identity continuity and association reliability \cite{lin2021trackletswitch,zhou2023ffattack,long2024papmot}. Visible--infrared attacks point toward modality-aware redundancy rather than RGB-only countermeasures \cite{tpamipatch,hu2024irblocks,long2025cdupatch,long2026thermally}.

In practice, this implies layered diagnostics rather than a single universal fix. Per-frame hardening can reduce obvious detector failures, temporal reasoning can flag suspicious switches or trajectory breaks, and cross-modal monitoring can use visible--thermal disagreement as a cue. Deployment-oriented testing remains necessary as well, because many apparent improvements disappear once camera pipelines, scale, or carrier realism change. The central issue is therefore not choosing one defense category, but understanding how several partial defenses interact inside the full surveillance pipeline.

\subsection{Open problems}

\paragraph{Persistent identity manipulation.}
Multi-object tracking attacks are still fewer in number than detector attacks, but they are often closer to real surveillance harm because they target identity continuity rather than one-frame visibility \cite{lin2021trackletswitch,zhou2023ffattack,pang2024blurmot,long2024papmot}. More work is needed on physically realizable attacks that create sustained ID corruption and on defenses that can detect such corruption before it spreads through a track history.

\paragraph{Cross-modal physical design.}
Visible--infrared evasion has been demonstrated, but the field still lacks a clear account of which physical attributes transfer across channels and which remain modality-specific \cite{tpamipatch,hu2024irblocks,hu2024ircurves,long2025cdupatch}. This is still a scientific question about sensing and carrier design, not just an engineering detail.

\paragraph{Wearability and activation.}
Wearable attacks are no longer limited to printed clothing. Triggered patches and thermally activated garments show that the attack surface includes state changes and user control \cite{zhu2023tpatch,long2026thermally}. Benchmarks should therefore model whether a carrier is always active, conditionally active, and easy or difficult to distinguish from ordinary apparel.

\paragraph{From attack demonstrations to system audits.}
Many of the most useful papers in this area do more than maximize one attack metric. They expose missing evaluation axes and operational assumptions. Future surveillance robustness research would benefit from more work that behaves like a system audit, tying together sensing modality, time, identity, and carrier realism rather than optimizing a single detector-oriented objective \cite{nguyen2023surveillance,long2024papmot,long2025cdupatch,long2026thermally}.

\section{Conclusion}
\label{sec:conclusion}

Physical adversarial attacks on surveillance systems are no longer well described by the older picture of a patch fooling a detector in a single RGB image. The literature now spans tracking, visible--infrared sensing, wearable carriers, and conditionally activated attacks. Taken together, these threads suggest that the relevant unit of analysis is the surveillance pipeline rather than the isolated model.

This leads to a simple conclusion. Stronger attack papers should report persistence over time, behavior across sensing channels, and the realism of the deployed carrier. Stronger defenses should be judged by the same criteria. If future benchmarks move in that direction, they will provide a more faithful picture of what robustness means in real surveillance environments.

{
    \small
    \bibliographystyle{ieeenat_fullname}
    \bibliography{main}
}

\end{document}